\documentclass{article}

\PassOptionsToPackage{round}{natbib}

\usepackage[main, final]{neurips_2026}

\usepackage[utf8]{inputenc} 
\usepackage[T1]{fontenc}    
\usepackage{hyperref}       
\usepackage{url}            
\usepackage{booktabs}       
\usepackage{amsfonts}       
\usepackage{amsmath}
\usepackage{nicefrac}       
\usepackage{microtype}      
\usepackage{xcolor}         
\usepackage{graphicx}
\usepackage{soul}
\usepackage{comment}

\usepackage[dvipsnames]{xcolor}

\title{Distilling LLM Feedback for Lean Theorem Proving}

\author{%
  Gaëtan Narozniak \\
  FAIR at Meta, Inria \\
  \texttt{gaetan@meta.com} \\
  \And
  Gérard Biau \\
  Sorbonne Université, \\
  Institut universitaire de France \\
  \And
  Rémi Munos \\
  FAIR at Meta \\
   \AND
  Ahmad Rammal \\
 FAIR at Meta, \\
 CERMICS École des Ponts ParisTech
   \And
   Pierre Marion \\
   Inria, ENS, PSL Research University \\
   \texttt{pierre.marion@inria.fr} \\
}

\begin{document}

\maketitle

\begin{abstract}
Post-training for reasoning models typically combines supervised fine-tuning with reinforcement learning from verifiable rewards, most commonly with GRPO. However, this algorithm suffers from sparse rewards, limited exploration, and mode collapse. Building upon recent works on self-distillation, we propose \emph{Feedback Distillation}, a training method where the model is trained to match, at the token level, its own distribution conditioned on privileged feedback produced by a language model. Feedback Distillation is on-policy, offers token-level supervision, and can inject external knowledge. Evaluating our method for Lean~4 theorem-proving, we find that Feedback Distillation maintains greater diversity in generated trajectories than GRPO, yielding higher policy entropy and better pass@$k$ scaling. The two methods are complementary: initializing GRPO from a Feedback Distillation checkpoint outperforms either method alone. All in all, our results suggest a promising avenue to improve post-training for complex reasoning.
\end{abstract}

\section{Introduction}

Recent advances in AI have demonstrated remarkable progress in complex reasoning tasks \citep{li2022competition,roziere2023code,yao2023tree,besta2024graph,rein2024gpqa}. Mathematical reasoning has emerged as a particularly valuable testbed for these approaches due to its challenging search spaces, clear correctness criteria, and rich problem structure ranging from formal theorem proving to informal problem-solving \citep{lightman2023lets,trinh2024solving,alphaproof2025olympiad,peyronnet2026lemmabench}. A fruitful strategy uses verifiable rewards, where deterministic verifiers (e.g., Lean proof checkers) evaluate model outputs. 
In this setting, state-of-the-art post-training pipelines combine supervised fine-tuning (SFT) on expert solutions (coming either from human experts or from a stronger LLM) with reinforcement learning (RL) to maximize the reward provided by the verifier \citep{kumar2025llm}. The RL algorithm is usually GRPO \citep{shao2024deepseekmath}, an algorithm that samples multiple outputs per prompt and optimizes the policy by contrasting their rewards. 

However, this SFT+GRPO paradigm has significant limitations. 
First, the SFT phase is limited by the difficulty and cost of generating the fine-tuning dataset. Furthermore, it is inherently \emph{off-policy}: the model is trained on externally generated demonstrations rather than on trajectories sampled from its own policy, a distribution mismatch that can cause catastrophic forgetting outside the fine-tuning distribution \citep[see, e.g.,][]{ross2010efficient,agarwal2024onpolicy,shenfeld2026rl,shenfeld2026selfdistillationenablescontinuallearning}. Moreover, recent evidence suggests GRPO does not amplify the diversity of model answers nor favor exploration during training, but rather collapses onto modes with high reward already present before RL training \citep{yue2025doesreinforcementlearningreally}. 
This is particularly problematic in formal mathematics where exploration of large libraries of lemmas (e.g., Lean's Mathlib, \citealp{mathlib2020}) is essential for success.
Finally, GRPO provides only trajectory-level rewards, which offer poor credit assignment, and no learning signal when all samples fail.

\begin{figure}[t]
    \centering
    \includegraphics[width=\linewidth]{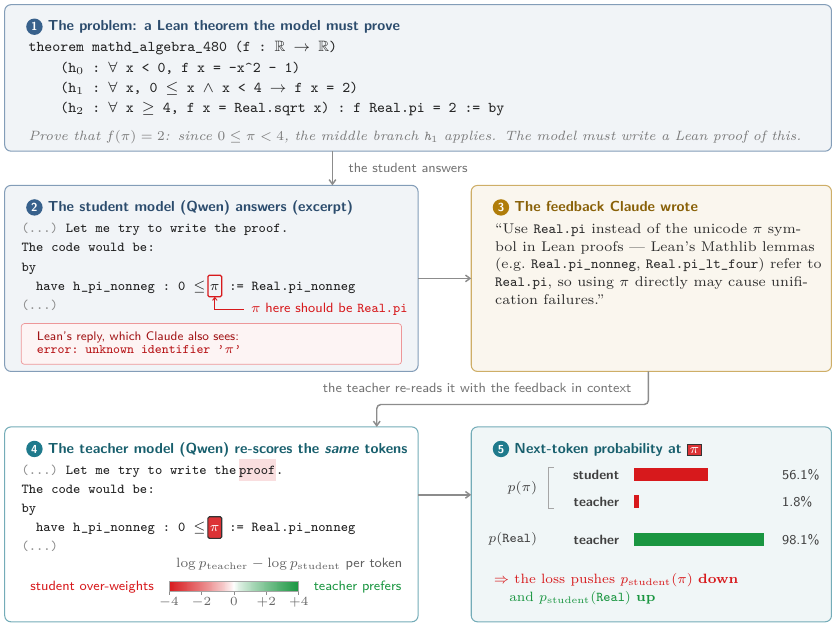}
    \caption{Example of Feedback Distillation on a Lean statement from the MiniF2F dataset \citep{zheng2022minif2fcrosssystembenchmarkformal}, with Qwen3-8B as the student. Claude provides feedback on the student's attempt. The student writes ``$\pi$'' where the teacher would have written ``\texttt{Real}'' with 98\% probability, reflecting Claude's feedback. Through the KL loss, the student is trained to increase the probability of \texttt{Real} at this position, and thus learns to write better Lean code.}
    \label{fig:token_logprob_viz}
\end{figure}

These interconnected challenges motivate exploration of alternative training mechanisms. 
A promising approach is the recently proposed self-distillation algorithm \citep{hubotter2026reinforcementlearningselfdistillation,shenfeld2026selfdistillationenablescontinuallearning,zhao2026selfdistilledreasoneronpolicyselfdistillation}, 
which trains the model to mimic its own outputs when conditioned on privileged information, distilling this information into the model's weights. This approach is attractive for two distinct reasons: the trajectories it trains on are sampled from the current policy, avoiding the distribution mismatch inherent to off-policy SFT; and the feedback-conditioned distribution is matched at the token level, providing fine-grained credit assignment rather than a reward distributed uniformly across entire trajectories (see example in Figure~\ref{fig:token_logprob_viz}). It thus combines the on-policy nature of GRPO with a supervision signal denser than that of SFT.
Moreover, it can apply beyond verifiable reward settings since it does not directly rely on verifier signals.

\paragraph{On-policy adaptation to a new environment.}
Proving theorems in Lean~4 is a capability our base models largely lack, so we view training on it as adapting the model to a new environment. Our focus is on whether that new ability can be acquired \emph{on-policy}, which has been shown to enable better continual learning and to preserve existing capabilities more effectively than off-policy training such as SFT \citep{agarwal2024onpolicy,shenfeld2026rl,shenfeld2026selfdistillationenablescontinuallearning}. Using a lagged version of the student's weights to update the teacher creates a feedback loop between the two models, which plays a crucial role in stabilizing the training process and getting a good performance.

\paragraph{Contributions.} In this work, we investigate the performance of self-distillation in the context of training LLMs for mathematics. More precisely,
\begin{itemize}
    \item We propose \emph{Feedback Distillation}, a variant of
    self-distillation in which the privileged information is
    feedback produced by a third-party LLM or a frozen copy of the model in response
    to the model’s attempt and the verifier’s output, rather than
    raw environment signals or ground-truth solutions 
    (Section~\ref{sec:feedback_distillation}).

    \item We evaluate Feedback Distillation in the Lean~4 formal
    mathematics environment (Section~\ref{sec:exploration}). We show that it maintains
    higher policy entropy than GRPO and achieves
    better pass@$k$ scaling, indicating greater output diversity.
    Furthermore, we show that the two methods are complementary: initializing
    GRPO from a Feedback Distillation checkpoint outperforms
    GRPO alone. After training on LeanWorkbook, Qwen3-8B achieves 23\% pass@1 test accuracy on MiniF2F with GRPO alone compared to 31\% with Feedback Distillation+GRPO.

    \item We provide a detailed analysis of Feedback Distillation: we compare with other sources of feedback, and discuss training instability and qualitative consequences of our training methodology, which can be seen as a form of on-policy knowledge transfer (Section~\ref{sec:analysis}). 

\end{itemize}

\section{Related work} \label{sec:related-works}

\paragraph{Self-distillation for LLMs.}
A recent line of work trains LLMs by minimizing the
divergence between a student and a teacher that shares the
same architecture but is conditioned on privileged information.
\citet{hubotter2026reinforcementlearningselfdistillation} use environment outputs as the privileged signal while
\citet{shenfeld2026selfdistillationenablescontinuallearning,zhao2026selfdistilledreasoneronpolicyselfdistillation}
use a ground-truth solution as the privileged information.
Our work extends this line of research by using feedback from a frozen copy of the model or from a third-party LLM. Furthermore, we perform a careful analysis of the properties of self-distillation, demonstrating in particular that it maintains greater trajectory diversity than GRPO, which translates into better test-time scaling when starting from a weak student in a difficult RL environment (formal mathematics).

\paragraph{RL for formal mathematics.}
State-of-the-art provers in formal mathematics usually follow a two-phase pipeline \citep{lin2025goedelproverv2scalingformaltheorem,ren2025deepseekproverv2advancingformalmathematical}:
supervised fine-tuning on synthetic proof data coming from the formalization of natural-language proofs, followed by
GRPO-based RL, often referred to as reinforcement learning with verifiable rewards (RLVR).
Recent progress on benchmarks such as MiniF2F \citep{zheng2022minif2fcrosssystembenchmarkformal} and PutnamBench \citep{tsoukalas2024putnambenchevaluatingneuraltheoremprovers} 
has increasingly come from test-time scaling rather than improved
training \citep[e.g.,][]{varambally2026hilbertrecursivelybuildingformal}. This scaling is achieved through agentic pipelines combining a trained prover with
a reasoning model and iterative interaction with the Lean
environment.
While these scaffolding strategies are orthogonal to our work,
they underscore the continued importance of strong base provers, whose training we aim to improve.

\section{Feedback Distillation} \label{sec:feedback_distillation}

Following the self-distillation methodology \citep{hubotter2026reinforcementlearningselfdistillation,shenfeld2026selfdistillationenablescontinuallearning,zhao2026selfdistilledreasoneronpolicyselfdistillation}, our training loss measures the divergence between two instances of the same model with independent weights, a \emph{student} and a \emph{teacher}, the latter receiving privileged information in its prompt. Both instances are initialized from the same pretrained LLM $\pi$.
We seek to post-train on a task for which we have a dataset $D$ of problems without solutions (e.g., Lean statements without proofs).

Specifically, given an attempted proof $y \sim \pi_\theta(\cdot \mid x)$ for a problem $x \sim D$, the teacher is given by
$ \pi_\mu(\cdot \mid x,\, F(y))$
where $F$ is a function that takes an attempted response $y$ and returns a textual feedback signal (the privileged information).
For consistency with prior work, we refer to it as the teacher model, though it would be more accurate to describe it as an augmented student.   

In our setting, the feedback $F$ is an LLM that analyzes the attempt $y$. Since the feedback may originate from a stronger model, we use the term \emph{Feedback Distillation}. The feedback model may also be a frozen copy of~$\pi$, which we sometimes refer to as Self-Feedback Distillation. We refer to Sections~\ref{sec:related-works} and~\ref{subsec:other-feedback} for discussion on other sources of feedback.
We give examples of feedback in Figure~\ref{fig:token_logprob_viz} and Appendix~\ref{app:feedback_examples}.

Following \citet{hubotter2026reinforcementlearningselfdistillation}, we update the teacher's weights via exponential moving average (EMA): every five gradient steps, we set $\mu \leftarrow \alpha\, \mu + (1-\alpha)\,\theta$, where $\alpha \in [0,1]$ is a hyperparameter. This hyperparameter has a critical impact on stability and performance (see Section~\ref{subsec:ema}).

Our loss is the per-token Kullback-Leibler divergence between the teacher and student distributions along trajectories sampled from the student, namely:
\[
\mathcal{L}_{\theta}
\;=\;
\mathbb{E}_{x \sim D,\; y \sim \pi_\theta(\cdot \mid x)}
\bigg[
\sum_{t=1}^{|y|}
\mathrm{KL}\!\left(
\pi_{\mu}(\cdot \mid x,F(y), y_{<t}) \;\|\; \pi_\theta(\cdot \mid x, y_{<t})
\right)
\bigg],
\]
where gradients are not propagated through the sampling of $y$, and $|y|$ denotes the length of the sequence~$y$. With $V$ the vocabulary, the per-token KL divergence is
\[
\mathrm{KL}\!\left(
\pi_{\mu}(\cdot \mid x,F(y), y_{<t}) \;\|\; \pi_\theta(\cdot \mid x, y_{<t})
\right)
\;=\;
\sum_{a \in V}
\pi_{\mu}(a \mid x,F(y), y_{<t})
\log \frac{
\pi_{\mu}(a \mid x,F(y), y_{<t})
}{
\pi_\theta(a \mid x, y_{<t})
}.
\]

Since the teacher's distribution does not depend on $\theta$, the KL divergence and the cross-entropy have the same gradient with respect to $\theta$ (see Appendix~\ref{app:loss_derivation}).
For a sampled problem $x \sim D$ and attempt $y \sim \pi_\theta(\cdot \mid x)$ with feedback $F(y)$, this yields the following equivalent loss, which we use in practice:
\[
\hat{\mathcal{L}}_\theta
\;=\;
- \sum_{t=1}^{|y|} \sum_{a \in V}
\pi_\mu(a \mid x,\, F(y),\, y_{<t})\;
\log \pi_\theta(a \mid x,\, y_{<t}).
\]

\paragraph{Top-$K$ truncation.}
To reduce computational cost, we truncate the sum over the vocabulary to the $K$ tokens with the highest probability under the teacher $\pi'_{\mu,y}$, using $K=25$ in practice.

\section{Capabilities of Feedback Distillation for Lean Theorem Proving} \label{sec:exploration}

In this section, we provide evidence that training with Feedback Distillation enables the model to learn efficiently, outperforming GRPO in performance and policy diversity.

\subsection{Lean environment}

Our experimental setting is the Lean~4 formal mathematics environment~\citep{10.1007/978-3-030-79876-5_37}.
It is a well-suited testbed for studying RL for complex reasoning
for several reasons: (i)~our base models (Qwen3-8B, \citealp{qwen3technicalreport}, and Qwen3.5-9B, \citealp{qwen35blog}) have little
proficiency in Lean~4, scoring respectively 2\% and 11\% on MiniF2F at
initialization in our setup, so any meaningful improvement requires discovering new
capabilities;
(ii)~the search space is vast, as the model must learn to invoke
appropriate tactics and reference lemmas from Mathlib, Lean's
main mathematical library comprising over 300{,}000 declarations;
(iii)~the Lean verifier provides ground-truth feedback on
whether a proof attempt is correct.
Results in this section are given for Qwen3-8B, and for Qwen3.5-9B in Section~\ref{sec:analysis}.

\paragraph{Tool interface.}
Tool use is a key component of state-of-the-art formal mathematics models~\citep{liu2026numina}, enabling iterative interaction with the proof assistant rather than single-shot generation. When the model produces a tool call, generation stops, the tool is executed, and its output is appended to the context before resuming generation.
During training, the loss is computed only on tokens generated by the model and not on tokens produced by the tools.
We equip the model with three tools to interact with the Lean environment.
\begin{itemize}
    \item \texttt{lean\_write\_file}: write a Lean source file;
    \item \texttt{lean\_check\_file}: compile a previously written file and return the compiler output (errors, warnings, or success);
    \item \texttt{rg\_in\_mathlib}: search Mathlib using regular expressions to find relevant lemmas.
\end{itemize}
We confirm in Appendix~\ref{app:tools_ablation} that tool use improves performance in our setting.
A problem is considered solved when the model writes a Lean file that compiles and proves the statement.

\subsection{Experimental setting}
\label{subsec:expe_setting}

We describe here the main ingredients of the experimental setting. Hyperparameters used can be found in Appendix~\ref{app:hyperparams}.

\paragraph{Datasets and generation.}
We train on a subset of 10{,}000 statements with known proofs drawn from LeanWorkbook~\citep{ying2025leanworkbooklargescalelean}.
For evaluation, we use a test set of LeanWorkbook comprising 256 statements as well as the test split of MiniF2F~\citep{zheng2022minif2fcrosssystembenchmarkformal}, a standard benchmark of 244 competition-level problems formalized in Lean.
Each attempt consists of a single generation of up to 8{,}192 tokens, within which the model may invoke multiple tool calls. All experiments are run on a single node equipped with 8× NVIDIA H200 GPUs. GRPO experiments take on average $550$s per training step, while Feedback Distillation requires $300$s per step. We use the \texttt{verl} codebase for the experiments \citep{sheng2024hybridflow} and the Kimina Lean Server \citep{santos2025kiminaleanservertechnical} to assess the correctness of Lean proofs.

\paragraph{Feedback model.}
We experiment with Claude Opus 4.6~\citep{anthropic_claude_opus_46_2026} as the feedback model (see Section~\ref{subsec:other-feedback} for discussion on sources of feedback).
The feedback model receives the formal statement $x$, the model's
attempt $y$ (including all tool calls and their
outputs), and the final Lean compiler output.
It is prompted to produce a few actionable directives of the form
\emph{``Do X''} or \emph{``Don't do X''}, identifying proof strategy
errors, flagging misuse of tactics or lemmas, or encouraging better
tool usage (see Appendix~\ref{app:feedback_prompt}).
The feedback is framed to be useful for a new attempt at
the same problem.
Examples of feedback are shown in Appendix~\ref{app:feedback_examples}.

\subsection{Results}    \label{subsec:results}

\begin{figure}[t]
    \centering
    \includegraphics[width=\linewidth]{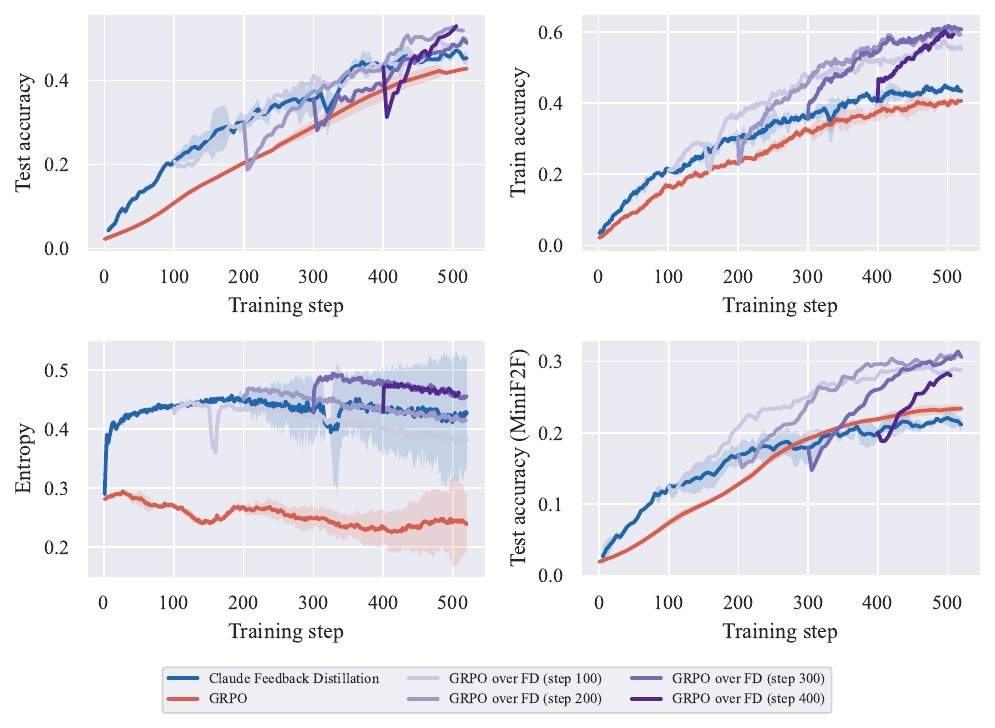}
    \caption{\textbf{Overview on Qwen3-8B.} LeanWorkbook test accuracy, train accuracy, policy entropy, and MiniF2F test accuracy for Claude Feedback Distillation, GRPO, and GRPO initialized from Feedback Distillation checkpoints at $100$, $200$, $300$ and $400$ steps. Curves are means over two seeds for GRPO and Claude Feedback Distillation, and shaded bands span $\pm 1$ standard deviation. GRPO over Feedback Distillation reaches the highest accuracy, suggesting that Feedback Distillation provides a good initialization for GRPO.}
    \label{fig:grpo_ckpt_qwen3_8b}
\end{figure}

We compare Feedback Distillation with GRPO, and also experiment with \emph{GRPO over Feedback
Distillation}, meaning that we take a checkpoint from Feedback Distillation and continue training
with GRPO.
Figure~\ref{fig:grpo_ckpt_qwen3_8b} gives an overview of this comparison for Qwen3-8B.
We observe that GRPO over Feedback Distillation reaches the highest accuracy, while Feedback Distillation maintains a policy entropy above that of GRPO.
Feedback Distillation is more compute efficient than GRPO, as it draws one sample per prompt where GRPO draws five; Table~\ref{tab:training_cost} in Appendix~\ref{app:compute_comparison} gives the full accounting.
Overall, \textbf{Feedback Distillation is able to learn efficiently while maintaining diversity in the generated trajectories}, a desirable property in light of recent findings that GRPO tends to reduce the diversity of model outputs during training \citep{dang2025assessing,yue2025doesreinforcementlearningreally,wu2026invisibleleashrlvrescape}.

\paragraph{GRPO over Feedback Distillation performs best.} We take checkpoints after $100$, $200$, $300$, and $400$ steps of Feedback
Distillation and continue training with GRPO.
As shown in Figure~\ref{fig:grpo_ckpt_qwen3_8b}, this combination outperforms GRPO
alone. 
This suggests that the two methods provide complementary benefits. 

\begin{figure}[h]
\centering
\includegraphics[width=0.9\linewidth]{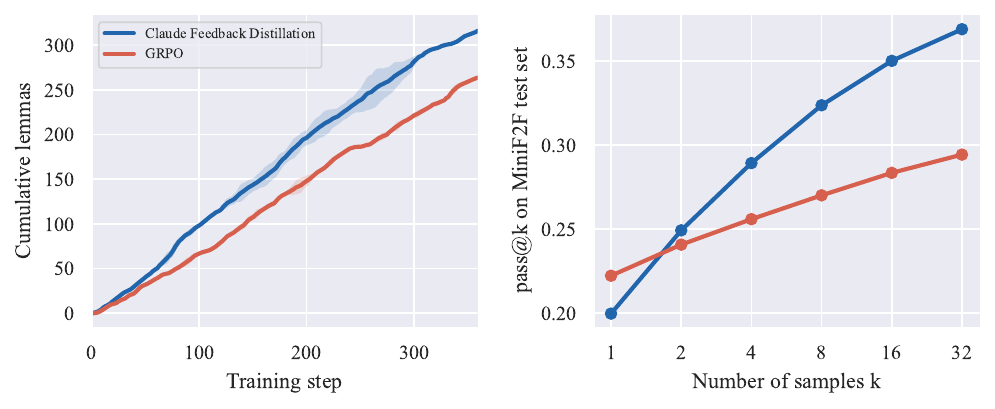}
\caption{(Left) Discovery of Mathlib lemmas: cumulative number of distinct lemmas used in correct proofs over the course of training. Feedback          
  Distillation leads to faster discovery of new lemmas from the Mathlib library. (Right) Pass@$k$ scaling for two  
  checkpoints at 400 steps of training with GRPO and Feedback Distillation (from the experiment in Figure~\ref{fig:grpo_ckpt_qwen3_8b}), where  
  both methods have similar pass@1 accuracy. Feedback Distillation scales better with $k$, confirming that its higher entropy translates into greater answer diversity.}              
\label{fig:passatk}                                   
\end{figure}

\paragraph{Number of discovered lemmas.}
A concrete manifestation of Feedback Distillation's greater trajectory diversity is the rate at which the model invokes new Mathlib lemmas in successful proofs over the course of training.
As shown in Figure~\ref{fig:passatk}, Feedback Distillation leads to a faster and more sustained increase in the number of distinct lemmas invoked, compared to GRPO.

\paragraph{Entropy and pass@$k$.} Feedback Distillation maintains significantly higher
  entropy than GRPO throughout training, indicating that the policy
  preserves more diversity (Figure~\ref{fig:grpo_ckpt_qwen3_8b}). 
  This translates into better test-time scaling: at checkpoints where GRPO and Feedback Distillation achieve similar pass@1, Feedback Distillation scales significantly better with   
  increasing $k$ (Figure~\ref{fig:passatk}), confirming that the higher entropy reflects answer diversity rather than uniform noise.

\paragraph{Sample efficiency.} Feedback Distillation draws a single sample per prompt where GRPO draws five, so at equal step count it has consumed five times fewer rollouts (see Appendix~\ref{app:hyperparams}). 

\section{Analysis of Feedback Distillation} \label{sec:analysis}

This section provides further analysis of Feedback Distillation. We begin by discussing other sources of feedback than LLM-generated advice based on a prior attempt, then the key role of EMA and its connection to training instability, before some qualitative comments.
\subsection{Other sources of feedback}  \label{subsec:other-feedback}

The experiments in this subsection use Qwen3.5-9B, as it more consistently picked up training signal from the different feedback sources than Qwen3-8B. We observe training instabilities for GRPO over Feedback Distillation on Qwen3.5-9B. Similar instabilities were observed in our setup when training the base Qwen3.5-9B model with GRPO. We expect that these instabilities could be mitigated by hyperparameter tuning, and do not affect the qualitative conclusions we draw in this subsection. We consider three alternative feedback sources: a ground truth proof \citep{shenfeld2026selfdistillationenablescontinuallearning,zhao2026selfdistilledreasoneronpolicyselfdistillation}, the raw Lean compiler output \citep{hubotter2026reinforcementlearningselfdistillation}, and the trained model itself, which we discuss separately below.
The ground truth feedback is a correct proof for the LeanWorkbook statement, formatted as \texttt{``Example of a correct solution:\{solution\}''}. In case of failure, the Lean output feedback is the raw Lean compiler output formatted as \texttt{``The following is feedback from your unsuccessful earlier attempt: \{environment\_output\}''}, and in case of success it is the solution found, with the same format as for the ground truth feedback. 

\begin{figure}[t]
    \centering
    \includegraphics[width=0.9\linewidth]{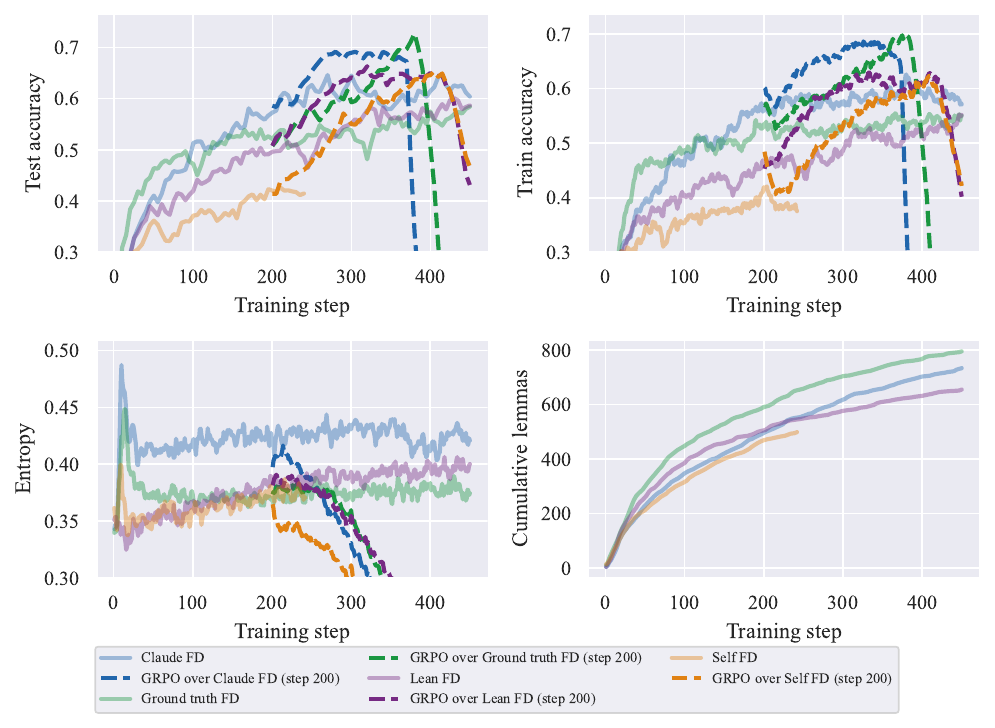}
    \caption{Feedback Distillation and GRPO over Feedback Distillation on Qwen3.5-9B for different feedback sources: Claude, a ground truth Lean solution, Lean compiler output, and Qwen itself (Self-Feedback Distillation). Claude and ground truth feedback sources are a better initialization for the GRPO training.}
    \label{fig:feedback_sources}
\end{figure}

A comparison of the performance is presented in Figure~\ref{fig:feedback_sources}. We observe that Claude and ground-truth as feedback sources outperform Lean output. This is expected, as both settings provide access to stronger sources of information. We hypothesize that the performance of Claude-based feedback could be further improved through better prompt design. In principle, it could even outperform both Lean output and ground truth, since Claude has access to compiler outputs and is sufficiently capable to generate correct solutions for LeanWorkbook problems. 

\paragraph{Self-Feedback Distillation.}
We consider an alternative source of feedback in which the feedback model is not a stronger external model, but the same model being trained, with frozen weights. We refer to this setting as \emph{Self-Feedback Distillation}. 
This setup does not introduce any additional source of information, and thus isolates the model’s ability to improve based on its own analysis of its generations.  
The performance does not yet match that of the other feedback sources.
We believe that further improvements may be achieved in this direction by refining the content and formulation of the feedback, pointing to promising directions for future work.

\subsection{EMA and training instability}   \label{subsec:ema}

Since the model learns from multiple feedback signals over the course of training, a frozen teacher would be problematic: knowledge acquired from earlier feedback would be penalized, since the teacher would not internalize this information and assign it low probability.
EMA updates of the teacher address this issue by letting it progressively track the student (Section~\ref{sec:feedback_distillation}). This echoes self-supervised learning, where EMA target networks are widely used to prevent the representational collapse arising from moving targets \citep[see, e.g.,][]{grill2020bootstrap,he2020momentum,assran2023self}.
The EMA interpolation parameter $\alpha \in [0,1]$ controls the tradeoff between adaptivity and stability. To assess this, we report in Figure~\ref{fig:lerp_sweep} a sweep over~$\alpha$ (see Appendix~\ref{app:ema_qwen35_9b} for the same sweep with Qwen3.5-9B). We find that more aggressive updates lead to faster learning early in training, while $\alpha=1$ avoids instability at the cost of slow learning. Nevertheless, the long-term instability of training even with large $\alpha$ suggests stabilization as an important area for future work.

\begin{figure}[t]
    \centering
    \includegraphics[width=0.9\linewidth]{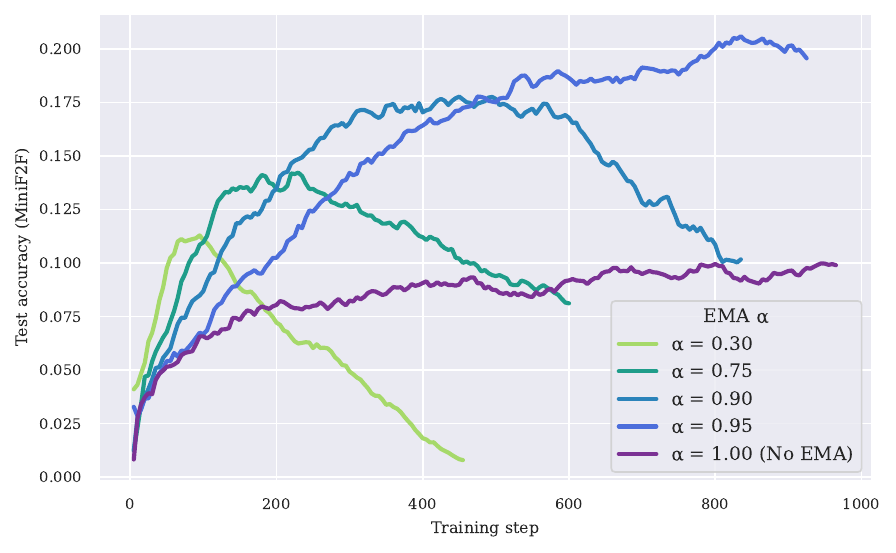}
    \caption{MiniF2F accuracy for Qwen3-8B during training for different values of the EMA interpolation parameter $\alpha$. More aggressive EMA updates (i.e., lower values of $\alpha$) accelerate early learning but result in increased instability. The case $\alpha = 1$ corresponds to freezing the teacher weights. We use $\alpha = 0.9$ for all the experiments.}
    \label{fig:lerp_sweep}
\end{figure}

\subsection{Qualitative comments on Feedback Distillation}  \label{subsec:qualitative}

We give below a few additional qualitative comments on Feedback Distillation.

\begin{figure}[t]
    \centering
    \includegraphics[width=\linewidth]{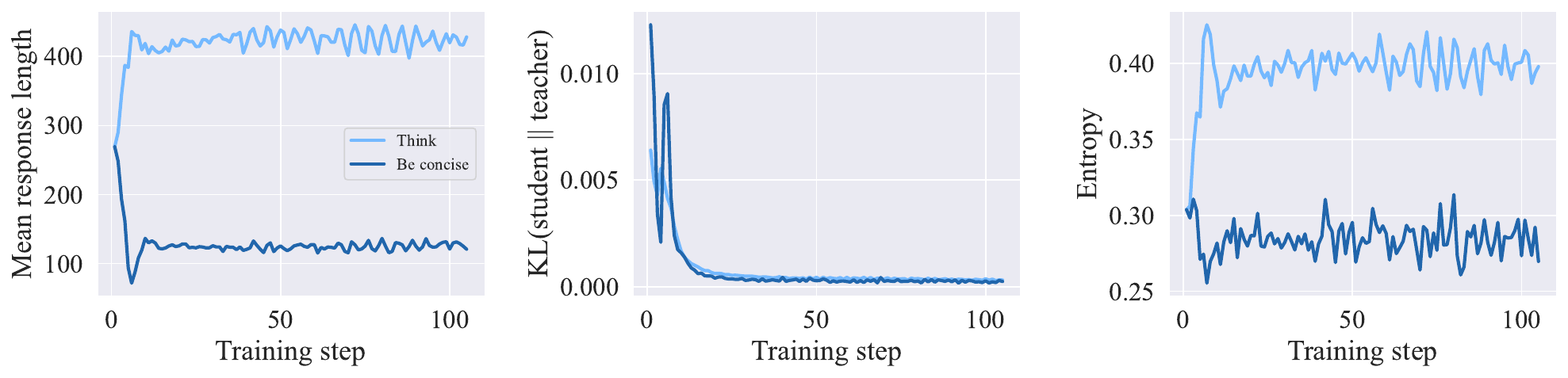}
    \caption{Effect of constant feedback on response length and policy entropy. ``Think'' feedback increases response length while ``Be concise'' decreases it. The KL divergence between student and teacher converges to zero, confirming that the feedback is internalized into the model's weights.}
    \label{fig:think_vs_concise}
\end{figure}

\paragraph{The feedback is internalized into the model's weights.} To verify this, we study a simplified setting where the
feedback $F(y)$ is a fixed string throughout training.
We consider two cases: ``Be concise'' or
``Think carefully before answering. Verify that the solution is
correct.''
In this experiment, the teacher is frozen.
As shown in Figure~\ref{fig:think_vs_concise}, the average response length evolves consistently with the feedback, and the KL divergence between student and teacher vanishes. This confirms that the student learns to match the teacher's behavior by distilling the privileged information into its weights: \textbf{after training, the student behaves as if the feedback were concatenated to every prompt}.

\paragraph{Feedback formulation.}
The formulation of the prompt sent to the feedback model to generate feedback plays a key role in training performance. Increasing the level of structure and guidance in this prompt (e.g., enforcing formatting constraints and requiring self-contained feedback) improves stability and final performance, while slowing early learning. We refer to Appendix~\ref{app:prompt_ablation} for a detailed analysis.

\paragraph{Epistemic verbalization.}
\citet{kim2026does} show that self-distillation with rich privileged information (e.g., a ground-truth solution) can suppress epistemic verbalization, i.e., the model's use of uncertainty words such as \texttt{wait}, \texttt{hmm}, \texttt{perhaps}, leading to degraded out-of-distribution performance.
We observe in Appendix~\ref{app:epistemic_words} that this suppression occurs significantly less often when the feedback is phrased as a critique, as in Feedback Distillation, suggesting that the information content of the conditioning signal can control this effect.

\paragraph{Feedback Distillation as on-policy knowledge transfer.}
When the feedback model is stronger than the student, Feedback Distillation acts as indirect knowledge distillation: the strong model provides information without needing to supply a full solution, unlike in standard distillation.
As discussed in the introduction, the process is also inherently on-policy, as the training signal is computed on the student's generations, a property shown to improve generalization and reduce catastrophic forgetting compared to off-policy methods such as SFT \citep{shenfeld2026selfdistillationenablescontinuallearning}. 
We provide in Appendix~\ref{app:fd_vs_sft} a comparison between using SFT and Feedback Distillation as an initialization for a GRPO run. 
Our results show that Feedback Distillation gets comparable results with an SFT-based pipeline while being on-policy. We believe this opens exciting research directions towards fully on-policy post-training of LLMs at scale.
A further practical advantage over standard on-policy distillation \citep{agarwal2024onpolicy} is that knowledge is transferred through natural-language feedback rather than logit matching, removing the need to host the feedback model or share its tokenizer.

\section{Conclusion}    \label{sec:discussion}

We view Feedback Distillation as a promising mechanism that addresses fundamental limitations of existing approaches by injecting external knowledge, maintaining diversity, and providing fine-grained credit assignment. More broadly, current LLMs appear to have sufficient reasoning capabilities to analyze their own successes and failures in grounded environments like formal mathematics, and Feedback Distillation seems a promising direction to internalize this ability into the model's weights. 
Our work opens several directions for future work. Training stability remains a challenge for Feedback Distillation; developing more robust stabilization mechanisms is a natural next step toward scaling to longer training runs. Additionally, while our results in Lean theorem proving are encouraging, extending Feedback Distillation to other formal and informal reasoning tasks could further demonstrate the generality of the approach.

\bibliographystyle{plainnat}
\bibliography{references}

@misc{yue2025doesreinforcementlearningreally,
      title={Does Reinforcement Learning Really Incentivize Reasoning Capacity in {LLM}s Beyond the Base Model?},
      author={Yang Yue and Zhiqi Chen and Rui Lu and Andrew Zhao and Zhaokai Wang and Yang Yue and Shiji Song and Gao Huang},
      year={2025},
      eprint={2504.13837},
      archivePrefix={arXiv},
      primaryClass={cs.AI}
}

@software{vonwerra2020trl,
  title   = {{TRL: Transformers Reinforcement Learning}},
  author  = {von Werra, Leandro and Belkada, Younes and Tunstall, Lewis and Beeching, Edward and Thrush, Tristan and Lambert, Nathan and Huang, Shengyi and Rasul, Kashif and Gallouédec, Quentin},
  license = {Apache-2.0},
  url     = {https://github.com/huggingface/trl},
  year    = {2020}
}

@inproceedings{shenfeld2026rl,
  title={Rl's razor: Why online reinforcement learning forgets less},
  author={Shenfeld, Idan and Pari, Jyothish and Agrawal, Pulkit},
  booktitle={International Conference on Learning Representations},
  volume={2026},
  pages={59839--59864},
  year={2026}
}

@misc{shenfeld2026selfdistillationenablescontinuallearning,
      title={Self-Distillation Enables Continual Learning}, 
      author={Idan Shenfeld and Mehul Damani and Jonas Hübotter and Pulkit Agrawal},
      year={2026},
      eprint={2601.19897},
      archivePrefix={arXiv},
      primaryClass={cs.LG}
}

@misc{hubotter2026reinforcementlearningselfdistillation,
      title={Reinforcement Learning via Self-Distillation}, 
      author={Jonas Hübotter and Frederike Lübeck and Lejs Behric and Anton Baumann and Marco Bagatella and Daniel Marta and Ido Hakimi and Idan Shenfeld and Thomas Kleine Buening and Carlos Guestrin and Andreas Krause},
      year={2026},
      eprint={2601.20802},
      archivePrefix={arXiv}
}

@misc{qwen3technicalreport,
      title={Qwen3 Technical Report}, 
      author={Qwen Team},
      year={2025},
      eprint={2505.09388},
      archivePrefix={arXiv},
      primaryClass={cs.CL}
}

@inproceedings{10.1007/978-3-030-79876-5_37,
  title = {The {L}ean 4 Theorem Prover and Programming Language},
  author = {de Moura, Leonardo and Ullrich, Sebastian},
  year = {2021},
  publisher = {Springer},
  address = {Berlin},
  booktitle = {Automated Deduction – CADE 28: 28th International Conference on Automated Deduction, Virtual Event, July 12–15, 2021, Proceedings},
  pages = {625–635},
  numpages = {11}
}

@inproceedings{mathlib2020,
  author    = {{The mathlib Community}},
  title     = {The {L}ean {M}athematical {L}ibrary},
  booktitle = {Proceedings of the 9th {ACM} {SIGPLAN} International Conference
               on Certified Programs and Proofs},
  series    = {CPP 2020},
  publisher = {ACM},
  address   = {New Orleans},
  year      = {2020}
}

@misc{ying2025leanworkbooklargescalelean,
      title={Lean Workbook: A large-scale {L}ean problem set formalized from natural language math problems}, 
      author={Huaiyuan Ying and Zijian Wu and Yihan Geng and Zheng Yuan and Dahua Lin and Kai Chen},
      year={2025},
      eprint={2406.03847},
      archivePrefix={arXiv},
      primaryClass={cs.CL}
}

@misc{zheng2022minif2fcrosssystembenchmarkformal,
      title={{MiniF2F}: A cross-system benchmark for formal {O}lympiad-level mathematics}, 
      author={Kunhao Zheng and Jesse Michael Han and Stanislas Polu},
      year={2022},
      eprint={2109.00110},
      archivePrefix={arXiv},
      primaryClass={cs.AI}
}

@online{anthropic_claude_opus_46_2026,
  author = {{Anthropic}},
  title  = {Introducing {C}laude Opus 4.6},
  year   = {2026},
  month  = feb,
  url    = {https://www.anthropic.com/news/claude-opus-4-6},
  note   = {Accessed: 2026-04-14}
}

@misc{zhao2026selfdistilledreasoneronpolicyselfdistillation,
      title={Self-Distilled Reasoner: On-Policy Self-Distillation for Large Language Models}, 
      author={Siyan Zhao and Zhihui Xie and Mengchen Liu and Jing Huang and Guan Pang and Feiyu Chen and Aditya Grover},
      year={2026},
      eprint={2601.18734},
      archivePrefix={arXiv},
      primaryClass={cs.LG}
}

@misc{lin2025goedelproverv2scalingformaltheorem,
      title={Goedel-Prover-{V}2: Scaling Formal Theorem Proving with Scaffolded Data Synthesis and Self-Correction}, 
      author={Yong Lin and Shange Tang and Bohan Lyu and Ziran Yang and Jui-Hui Chung and Haoyu Zhao and Lai Jiang and Yihan Geng and Jiawei Ge and Jingruo Sun and Jiayun Wu and Jiri Gesi and Ximing Lu and David Acuna and Kaiyu Yang and Hongzhou Lin and Yejin Choi and Danqi Chen and Sanjeev Arora and Chi Jin},
      year={2025},
      eprint={2508.03613},
      archivePrefix={arXiv},
      primaryClass={cs.LG}
}

@misc{ren2025deepseekproverv2advancingformalmathematical,
      title={DeepSeek-Prover-{V}2: Advancing Formal Mathematical Reasoning via Reinforcement Learning for Subgoal Decomposition}, 
      author={Z. Z. Ren and Zhihong Shao and Junxiao Song and Huajian Xin and Haocheng Wang and Wanjia Zhao and Liyue Zhang and Zhe Fu and Qihao Zhu and Dejian Yang and Z. F. Wu and Zhibin Gou and Shirong Ma and Hongxuan Tang and Yuxuan Liu and Wenjun Gao and Daya Guo and Chong Ruan},
      year={2025},
      eprint={2504.21801},
      archivePrefix={arXiv},
      primaryClass={cs.CL}
}

@misc{varambally2026hilbertrecursivelybuildingformal,
      title={Hilbert: Recursively Building Formal Proofs with Informal Reasoning}, 
      author={Sumanth Varambally and Thomas Voice and Yanchao Sun and Zhifeng Chen and Rose Yu and Ke Ye},
      year={2026},
      eprint={2509.22819},
      archivePrefix={arXiv},
      primaryClass={cs.AI}
}

@misc{wu2026invisibleleashrlvrescape,
      title={The Invisible Leash: Why {RLVR} May or May Not Escape Its Origin}, 
      author={Fang Wu and Weihao Xuan and Ximing Lu and Mingjie Liu and Yi Dong and Zaid Harchaoui and Yejin Choi},
      year={2026},
      eprint={2507.14843},
      archivePrefix={arXiv},
      primaryClass={cs.LG}
}

@inproceedings{
dang2025assessing,
title={Assessing Diversity Collapse in Reasoning},
author={Xingyu Dang and Christina Baek and J Zico Kolter and Aditi Raghunathan},
booktitle={Scaling Self-Improving Foundation Models without Human Supervision},
year={2025}
}

@InProceedings{ross2010efficient,
  title = 	 {Efficient Reductions for Imitation Learning},
  author = 	 {Ross, Stephane and Bagnell, Drew},
  booktitle = 	 {Proceedings of the Thirteenth International Conference on Artificial Intelligence and Statistics},
  pages = 	 {661--668},
  year = 	 {2010},
  volume = 	 {9}
}

@inproceedings{agarwal2024onpolicy,
title={On-Policy Distillation of Language Models: Learning from Self-Generated Mistakes},
author={Rishabh Agarwal and Nino Vieillard and Yongchao Zhou and Piotr Stanczyk and Sabela Ramos Garea and Matthieu Geist and Olivier Bachem},
booktitle={The Twelfth International Conference on Learning Representations},
year={2024}
}

@inproceedings{tsoukalas2024putnambenchevaluatingneuraltheoremprovers,
  title={Putnam{B}ench: Evaluating neural theorem-provers on the putnam mathematical competition},
  author={Tsoukalas, George and Lee, Jasper and Jennings, John and Xin, Jimmy and Ding, Michelle and Jennings, Michael and Thakur, Amitayush and Chaudhuri, Swarat},
  booktitle={Advances in Neural Information Processing Systems},
  volume={37},
  pages={11545--11569},
  year={2024},
  publisher = {Curran Associates, Inc.}
}

@article{alphaproof2025olympiad,
  title={Olympiad-level formal mathematical reasoning with reinforcement learning},
  author={Hubert, Thomas and Mehta, Rishi and Sartran, Laurent and Horv{\'a}th, Mikl{\'o}s Z and {\v{Z}}u{\v{z}}i{\'c}, Goran and Wieser, Eric and Huang, Aja and Schrittwieser, Julian and Schroecker, Yannick and Masoom, Hussain and others},
  journal={Nature},
  pages={1--3},
  year={2025},
  publisher={Nature Publishing Group UK London}
}

@article{kumar2025llm,
  title={{LLM} post-training: A deep dive into reasoning large language models},
  author={Kumar, Komal and Ashraf, Tajamul and Thawakar, Omkar and Anwer, Rao Muhammad and Cholakkal, Hisham and Shah, Mubarak and Yang, Ming-Hsuan and Torr, Phillip HS and Khan, Fahad Shahbaz and Khan, Salman},
  journal={arXiv preprint arXiv:2502.21321},
  year={2025}
}

@article{shao2024deepseekmath,
  title={Deepseekmath: Pushing the limits of mathematical reasoning in open language models},
  author={Shao, Zhihong and Wang, Peiyi and Zhu, Qihao and Xu, Runxin and Song, Junxiao and Bi, Xiao and Zhang, Haowei and Zhang, Mingchuan and Li, YK and others},
  journal={arXiv preprint arXiv:2402.03300},
  year={2024}
}

@article{peyronnet2026lemmabench,
  title={LemmaBench: A Live, Research-Level Benchmark to Evaluate {LLM} Capabilities in Mathematics},
  author={Peyronnet, Antoine and Gloeckle, Fabian and Hayat, Amaury},
  journal={arXiv preprint arXiv:2602.24173},
  year={2026}
}

@inproceedings{grill2020bootstrap,
  title={Bootstrap your own latent: A new approach to self-supervised learning},
  author={Grill, Jean-Bastien and Strub, Florian and Altch{\'e}, Florent and Tallec, Corentin and Richemond, Pierre H and Buchatskaya, Elena and Doersch, Carl and Pires, Bernardo Avila and Guo, Zhaohan Daniel and Azar, Mohammad Gheshlaghi and others},
  booktitle={Advances in Neural Information Processing Systems},
  volume={33},
  pages={21271--21284},
  year={2020},
   publisher = {Curran Associates, Inc.}
}

@inproceedings{he2020momentum,
  title={Momentum contrast for unsupervised visual representation learning},
  author={He, Kaiming and Fan, Haoqi and Wu, Yuxin and Xie, Saining and Girshick, Ross},
  booktitle={Proceedings of the IEEE/CVF Conference on Computer Vision and Pattern Recognition},
  pages={9729--9738},
  year={2020}
}

@inproceedings{assran2023self,
  title={Self-supervised learning from images with a joint-embedding predictive architecture},
  author={Assran, Mahmoud and Duval, Quentin and Misra, Ishan and Bojanowski, Piotr and Vincent, Pascal and Rabbat, Michael and LeCun, Yann and Ballas, Nicolas},
  booktitle={Proceedings of the IEEE/CVF Conference on Computer Vision and Pattern Recognition},
  pages={15619--15629},
  year={2023}
}

@article{lightman2023lets,
  title={Let's Verify Step by Step},
  author={Lightman, Hunter and Kosaraju, Vineet and Burda, Yura and Edwards, Harri and Baker, Bowen and Lee, Teddy and Leike, Jan and Schulman, John and Sutskever, Ilya and Cobbe, Karl},
  journal={arXiv preprint arXiv:2305.20050},
  year={2023}
}

@article{trinh2024solving,
  title={Solving olympiad geometry without human demonstrations},
  author={Trinh, Trieu H. and Wu, Yuhuai and Le, Quoc V. and He, He and Luong, Thang},
  journal={Nature},
  volume={625},
  number={7995},
  pages={476--482},
  year={2024},
  publisher={Nature Publishing Group}
}

@article{li2022competition,
  title={Competition-level code generation with AlphaCode},
  author={Li, Yujia and Choi, David and Chung, Junyoung and Kushman, Nate and Schrittwieser, Julian and Leblond, R{\'e}mi and Eccles, Tom and Keeling, James and Gimeno, Felix and Dal Lago, Agustin and others},
  journal={Science},
  volume={378},
  number={6624},
  pages={1092--1097},
  year={2022},
  publisher={American Association for the Advancement of Science}
}

@article{roziere2023code,
  title={Code llama: Open foundation models for code},
  author={Rozi{\`e}re, Baptiste and Gehring, Jonas and Gloeckle, Fabian and Sootla, Sten and Gat, Itai and Tan, Xiaoqing Ellen and Adi, Yossi and Liu, Jingyu and Sauvestre, Romain and Remez, Tal and Rapin, J{\'e}r{\'e}my and others},
  journal={arXiv preprint arXiv:2308.12950},
  year={2023}
}

@inproceedings{rein2024gpqa,
title={{GPQA}: A Graduate-Level Google-Proof Q\&A Benchmark},
author={David Rein and Betty Li Hou and Asa Cooper Stickland and Jackson Petty and Richard Yuanzhe Pang and Julien Dirani and Julian Michael and Samuel R. Bowman},
booktitle={First Conference on Language Modeling},
year={2024},
}

@article{yao2023tree,
  title={Tree of thoughts: Deliberate problem solving with large language models},
  author={Yao, Shunyu and Yu, Dian and Zhao, Jeffrey and Shafran, Izhak and Griffiths, Tom and Cao, Yuan and Narasimhan, Karthik},
  journal={Advances in neural information processing systems},
  volume={36},
  pages={11809--11822},
  year={2023}
}

@article{besta2024graph,
  title={Graph of Thoughts: Solving Elaborate Problems with Large Language Models},
  author={Besta, Maciej and Blach, Nils and Kubicek, Ales and Gerstenberger, Robert and Podstawski, Michal and Gianinazzi, Lukas and Gajda, Joanna and Lehmann, Tomasz and Niewiadomski, Hubert and Nyczyk, Piotr and others},
  journal={Proceedings of the AAAI Conference on Artificial Intelligence},
  volume={38},
  number={16},
  pages={17682--17690},
  year={2024}
}

@article{kim2026does,
  title={Why Does Self-Distillation (Sometimes) Degrade the Reasoning Capability of LLMs?},
  author={Kim, Jeonghye and Luo, Xufang and Kim, Minbeom and Lee, Sangmook and Kim, Dohyung and Jeon, Jiwon and Li, Dongsheng and Yang, Yuqing},
  journal={arXiv preprint arXiv:2603.24472},
  year={2026}
}

@article{liu2026numina,
  title={Numina-Lean-Agent: An Open and General Agentic Reasoning System for Formal Mathematics},
  author={Liu, Junqi and Zhou, Zihao and Zhu, Zekai and Santos, Marco Dos and He, Weikun and Liu, Jiawei and Wang, Ran and Xie, Yunzhou and Zhao, Junqiao and Wang, Qiufeng and others},
  journal={arXiv preprint arXiv:2601.14027},
  year={2026}
}

@article{kim2026understanding,
  title={Understanding Reasoning in LLMs through Strategic Information Allocation under Uncertainty},
  author={Kim, Jeonghye and Luo, Xufang and Kim, Minbeom and Lee, Sangmook and Li, Dongsheng and Yang, Yuqing},
  journal={arXiv preprint arXiv:2603.15500},
  year={2026}
}

@misc{qwen35blog,
    title = {Qwen3.5: Accelerating Productivity with Native Multimodal Agents},
    url = {https://qwen.ai/blog?id=qwen3.5},
    author = {Qwen Team},
    month = {February},
    year = {2026}
}

@article{sheng2024hybridflow,
  title   = {HybridFlow: A Flexible and Efficient RLHF Framework},
  author  = {Guangming Sheng and Chi Zhang and Zilingfeng Ye and Xibin Wu and Wang Zhang and Ru Zhang and Yanghua Peng and Haibin Lin and Chuan Wu},
  year    = {2024},
  journal = {arXiv preprint arXiv: 2409.19256}
}

@misc{santos2025kiminaleanservertechnical,
      title={Kimina Lean Server: Technical Report}, 
      author={Marco Dos Santos and Haiming Wang and Hugues de Saxcé and Ran Wang and Mantas Baksys and Mert Unsal and Junqi Liu and Zhengying Liu and Jia Li},
      year={2025},
      eprint={2504.21230},
      archivePrefix={arXiv},
      primaryClass={cs.LO},
      url={https://arxiv.org/abs/2504.21230}, 
}

\begin{ack}
  We are grateful to Julia Kempe for her help and valuable advice throughout the writing of this article.
\end{ack}


\clearpage
\appendix

\section{Setup and Implementation}

\subsection{Training hyperparameters}
\label{app:hyperparams}

Table~\ref{tab:hyperparams} lists the hyperparameters used in all Feedback Distillation and GRPO experiments reported in the paper.

\begin{table}[h]
    \centering
    \caption{Training hyperparameters for Feedback Distillation and GRPO experiments.}
    \label{tab:hyperparams}
    \vspace{0.5em}
    \begin{tabular}{lcc}
	\toprule
	Hyperparameter & Feedback Distillation & GRPO \\
	\midrule
	Learning rate & $1 \times 10^{-5}$ & $1 \times 10^{-7}$ \\
	Prompts per batch & 128 & 128 \\
	Samples per prompt & 1 & 5 \\
	Effective batch size & 128 & 640 \\
	EMA coefficient $\alpha$ & 0.9 & --- \\
	Top-$K$ truncation & 25 & --- \\
	KL regularization coeff. & --- & 0.001 \\
	Max response length & 8192 & 8192 \\
	Max prompt length & 2048 & 2048 \\
	  Linear learning rate warmup & 50 steps & 50 steps \\
	\bottomrule
    \end{tabular}
\end{table}

\subsection{Tool use ablation}
\label{app:tools_ablation}

\begin{figure}[h]
    \centering
    \includegraphics[width=\linewidth]{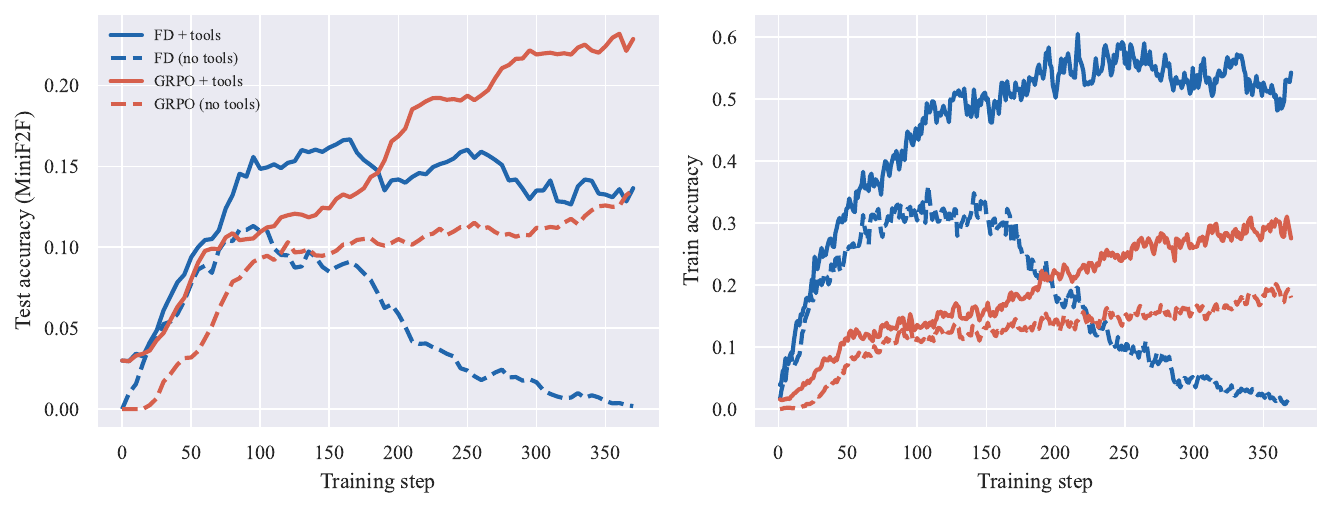}
    \caption{Qwen3-8B MiniF2F accuracy and training reward for Feedback Distillation and GRPO, with and without tool access. Tool use improves performance for both methods.}
    \label{fig:tools_vs_no_tools}
\end{figure}

Figure~\ref{fig:tools_vs_no_tools} compares training with and without tool access for both Claude Feedback Distillation and GRPO with Qwen3-8B.
In both cases, tool use leads to higher MiniF2F accuracy and better training accuracy, confirming that iterative interaction with the Lean compiler is a driver of performance~\citep{liu2026numina}.

\subsection{Feedback prompt}
\label{app:feedback_prompt}

The following prompt is given to the feedback model (Claude Opus 4.6) along with the formal statement $x$, the model's attempt $y$ (including tool calls and their outputs), and the final Lean compiler output.

\begin{quote}
    \small
    You are an expert feedback tutor. Given a problem and a student's attempt, write concise, actionable pieces of advice that will provide general feedback from the student's experience. Write between 1 and 3 numbered pieces of feedback about errors, things that were useful, or things that were not. The feedback must be self-contained and useful to someone who does not have access to the student's answer. It must be framed as orders, like ``Do X'' or ``Don't do X'', without mentioning the student. Each piece of feedback must be between 10 and 50 words. The goal is not to comment on the performance of the student, but to provide useful feedback for someone who might want to try the same problem in the future. Do not use ``you'', ``your'', or the second person in general. It must be useful to someone who does not know there is a student. It must encourage the usage of tools. Do not output anything other than the feedback — no introduction, directly the 1.\ of the first piece of advice. If the answer seems truncated at the end, it may be because the student reached the token limit; in this case, order to be more concise.
\end{quote}

\subsection{Compute comparison between GRPO and Feedback Distillation}
\label{app:compute_comparison}

Table~\ref{tab:training_cost} reports the compute spent by Claude Feedback Distillation and by GRPO to reach step $200$.
Feedback Distillation draws a single sample per prompt where GRPO draws five, so it reaches the same step count having generated $5\times$ fewer trajectories in $1.8\times$ less wall-clock time.
At current prices this is roughly \$250 of GPU compute against \$440 for GRPO, but Feedback Distillation additionally queries an external feedback model, at approximately \$1{,}100, so its saving in compute is more than offset by the API cost.
We give these figures as a point of information only: they track API pricing, which keeps falling as cheaper frontier models of comparable quality become available.

\begin{table}[h]
    \centering
    \caption{Training cost of Claude Feedback Distillation and GRPO, accumulated over the first $200$ training steps. Feedback Distillation reaches step $200$ with $5\times$ fewer rollouts and $1.8\times$ less wall-clock. }
    \label{tab:training_cost}
    \vspace{0.5em}
    \begin{tabular}{lccc}
	\toprule
	Metric (cumulative through step 200) & Claude FD & GRPO & GRPO\,/\,FD \\
	\midrule
	Rollouts per training step & 128 & 640 & $5.0\times$ \\
	Total rollouts & 25{,}600 & 128{,}000 & $5.0\times$ \\
	\addlinespace
	Wall-clock (h) & 15.85 & 27.74 & $1.8\times$ \\
	\quad Generation (h) & 5.15 & 11.39 & $2.2\times$ \\
	\quad Feedback (h) & 1.39 & 0 & --- \\
	\addlinespace
	Generated response tokens (M) & 137.1 & 801.0 & $5.8\times$ \\
	Lean tool calls (k) & 423.6 & 2{,}064.5 & $4.9\times$ \\
	\quad Mean tool calls per trajectory & 16.5 & 16.1 & $1.0\times$ \\
	\addlinespace
	Feedback-model calls & 25{,}600 & 0 & --- \\
	Feedback input tokens (M) & $\approx 168.2$ & 0 & --- \\
	Feedback output tokens (M) & $\approx 7.68$ & 0 & --- \\
	\bottomrule
    \end{tabular}
\end{table}

\section{Loss derivation}
\label{app:loss_derivation}

Since gradients are not propagated through the sampling of $y$ (stop-gradient), the gradient of $\mathcal{L}_\theta$ with respect to $\theta$ reduces to differentiating only through $\pi_\theta$ inside the KL:
\[
    \nabla_\theta \mathcal{L}_\theta
    \;=\;
    - \mathbb{E}_{x \sim D,\; y \sim \pi_\theta(\cdot \mid x)}
    \left[
	\sum_{t=1}^{|y|} \sum_{a \in V}
	\pi'_{\mu,y}(a \mid x, y_{<t})\;
	\nabla_\theta \log \pi_\theta(a \mid x, y_{<t})
	\right].
\]
Since the entropy of the teacher $H(\pi'_{\mu,y})$ does not depend on $\theta$, the KL divergence and the cross-entropy have the same gradient with respect to $\theta$. Expanding $\pi'_{\mu,y}$ by its definition, we obtain the gradient-equivalent loss used in practice:
\[
    \hat{\mathcal{L}}_\theta
    \;=\;
    - \sum_{t=1}^{|y|} \sum_{a \in V}
    \pi_\mu(a \mid x,\, F(y),\, y_{<t})\;
    \log \pi_\theta(a \mid x,\, y_{<t}).
\]

\section{Additional Results and Analysis}

\subsection{EMA sweep for Qwen3.5-9B}
\label{app:ema_qwen35_9b}

\begin{figure}[h]
    \centering
    \includegraphics[width=\linewidth]{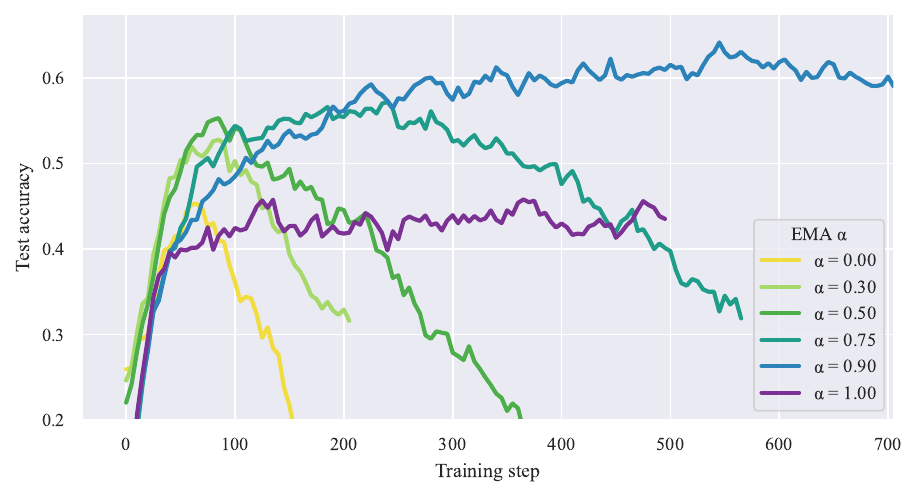}
    \caption{LeanWorkbook test accuracy for Qwen3.5-9B during training for different values of the EMA interpolation parameter $\alpha$. A more aggressive EMA (i.e., a lower value of $\alpha$) leads to faster training but also more instability. $\alpha = 1$ is the same as freezing the teacher weights.}
    \label{fig:lerp_sweep_qwen35}
\end{figure}

Consistent with the observations in Figure~\ref{fig:lerp_sweep} for Qwen3-8B, we find in Figure~\ref{fig:lerp_sweep_qwen35} that the same trend holds for Qwen3.5-9B: more aggressive EMA updates lead to faster early learning, but induce greater instability and ultimately limit peak performance.

\subsection{Comparison between Feedback Distillation and SFT}
\label{app:fd_vs_sft}

\begin{figure}[h]
    \centering
    \includegraphics[width=\linewidth]{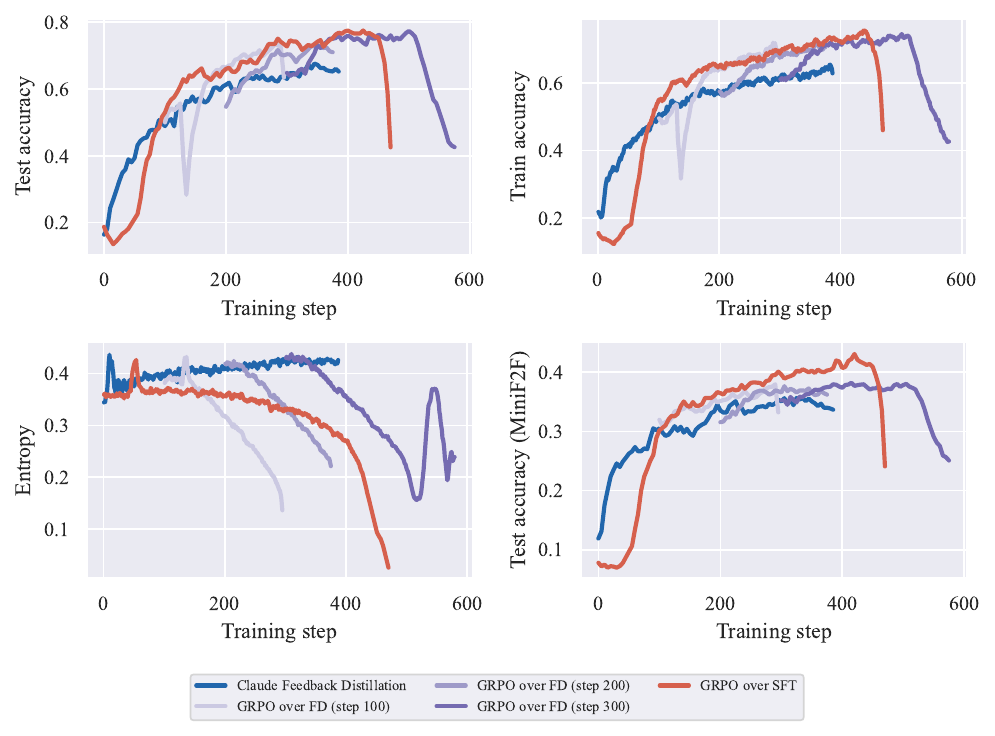}
    \caption{\textbf{GRPO over SFT versus Feedback Distillation on Qwen3.5-9B.}
    LeanWorkbook test accuracy, train accuracy, policy entropy and MiniF2F test
    accuracy, for Feedback Distillation, GRPO resumed from Feedback Distillation
    checkpoints at $100$, $200$ and $300$ steps, and GRPO resumed from a
    SFT checkpoint.}
    \label{fig:fd_vs_sft}
\end{figure}

The GRPO over SFT curve in Figure~\ref{fig:fd_vs_sft}
starts from a supervised-finetuned Qwen3.5-9B rather than from a Feedback
Distillation checkpoint, so that the two initializations can be compared under
the same GRPO recipe.

The SFT data was made by prompting Claude to prove Lean Workbook theorems with access to Lean tools, recording both its reasoning and its final proof. $9{,}494$
training examples are thus built, with a median of roughly $1{,}900$ characters of reasoning
and $170$ characters of proof. 

Training used TRL's \texttt{SFTTrainer} \citep{vonwerra2020trl} on a single $8\times$H200 node with
DeepSpeed ZeRO-2; Table~\ref{tab:sft_hyperparams} lists the hyperparameters. At
an effective batch size of $64$, one epoch is $148$ optimizer steps, so the
checkpoint at step 200 we branch from corresponds to $1.35$ epochs, or
$12{,}800$ examples seen.

The goal here is to compare Feedback Distillation with SFT as a first training phase before a GRPO training. Both methods get similar results, and are a way to distill Claude Lean knowledge, either off-policy by training on its generated proofs, or on-policy by using its feedback only (and not whole solution) as privileged information in Feedback Distillation.
We see Feedback Distillation as a different and reasonable way to distill knowledge in an on-policy way.

\begin{table}[h]
    \centering
    \caption{Hyperparameters for the supervised finetuning that initializes the
    GRPO over SFT run.}
    \label{tab:sft_hyperparams}
    \vspace{0.5em}
    \begin{tabular}{ll}
      \toprule
      Hyperparameter & Value \\
      \midrule
      Base model & Qwen3.5-9B \\
      Learning rate & $1 \times 10^{-5}$ \\
      Schedule & cosine, warmup ratio $0.03$ \\
      Optimizer & AdamW, $\beta = (0.9, 0.999)$, weight decay $0$ \\
      Gradient clipping & $1.0$ \\
      Effective batch size & 64 ($2 \times 4$ accumulation $\times\, 8$ GPUs) \\
      Epochs & 1.35 \\
      Max sequence length & 8192 \\
      \bottomrule
    \end{tabular}
\end{table}

\subsection{Feedback formulation}
\label{app:prompt_ablation}

We investigate how the formulation of feedback affects training performance. To this end, we evaluate four different prompt styles for Claude Opus 4.6 as the feedback model, while keeping all other hyperparameters fixed. The prompts are ordered from (1) to (4) by increasing level of structure and guidance, with each successive prompt introducing additional instructions: (1) a minimal prompt requesting feedback, (2) the addition of formatting constraints (e.g., length and number of feedback items), (3) a requirement that the feedback be self-contained and useful to a reader without access to the student’s answer, and (4) additional guidance encouraging the use of tools and limiting generation length in case of truncation. We use Prompt~(4) in all experiments.

\paragraph{Prompt templates.}
We provide below the four prompt formulations used in this study:

\begin{itemize}
    \item \textbf{Prompt (1): Minimal feedback request}
    \begin{quote}
    \small
    You are an expert feedback tutor. Given a problem and a student's attempt, write concise, actionable pieces of advice based on what the student did.
    \end{quote}

    \item \textbf{Prompt (2): + Formatting constraints}
    \begin{quote}
    \small
    You are an expert feedback tutor. Given a problem and a student's attempt, write concise, actionable pieces of advice based on what the student did. Write between 1 and 3 numbered feedbacks about errors, things that were useful, or things that were not. They must not be too long — between 10 and 50 words each. Do not output anything else than the feedbacks, no introduction, directly the "1." of the first advice.
    \end{quote}

    \item \textbf{Prompt (3): + Self-contained feedback requirement}
    \begin{quote}
    \small
    You are an expert feedback tutor. Given a problem and a student's attempt, write concise, actionable pieces of advice based on what the student did. Write between 1 and 3 numbered feedbacks about errors, things that were useful, or things that were not. They must not be too long — between 10 and 50 words each. Do not output anything else than the feedbacks, no introduction, directly the "1." of the first advice. The feedback must be self-contained and useful to someone who does not have access to the student's answer. Don't use "you", "your", or the second person in general — it must be useful to someone who doesn't know there is a student. It must be framed as orders, like "Do x" or "Don't do x", without mentioning the student. The goal is not to comment on the performance of the student, but to provide useful guidance for someone who might want to try the same problem in the future.
    \end{quote}

    \item \textbf{Prompt (4): + Tool use and truncation guidance}
    \begin{quote}
    \small
    You are an expert feedback tutor. Given a problem and a student's attempt, write concise, actionable pieces of advice that will provide general feedback from the student's experience. Write between 1 and 3 numbered pieces of feedback about errors, things that were useful, or things that were not. The feedback must be self-contained and useful to someone who does not have access to the student's answer. It must be framed as orders, like ``Do X'' or ``Don't do X'', without mentioning the student. Each piece of feedback must be between 10 and 50 words. The goal is not to comment on the performance of the student, but to provide useful feedback for someone who might want to try the same problem in the future. Do not use ``you'', ``your'', or the second person in general. It must be useful to someone who does not know there is a student. It must encourage the usage of tools. Do not output anything other than the feedback — no introduction, directly the 1.\ of the first piece of advice. If the answer seems truncated at the end, it may be because the student reached the token limit; in this case, order to be more concise.
    \end{quote}
\end{itemize}

\begin{figure}[h]
    \centering
    \includegraphics[width=\linewidth]{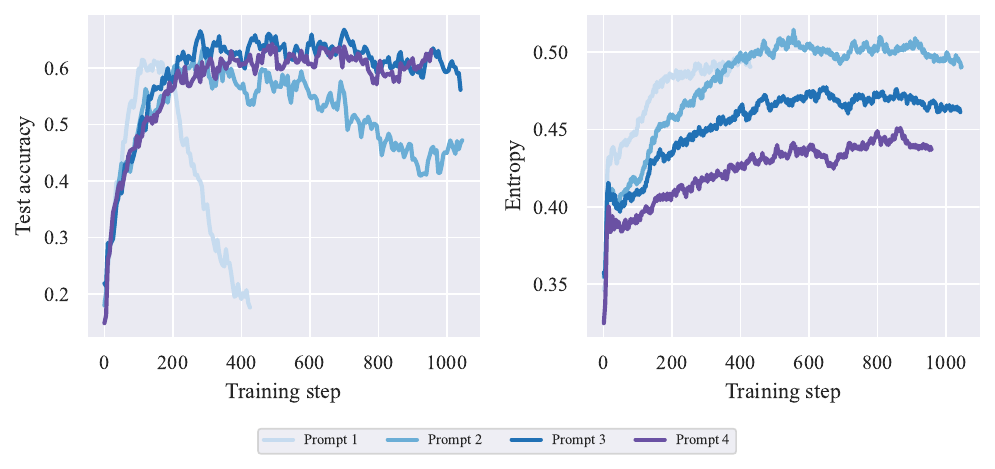}
    \caption{Impact of feedback prompt formulation on training performance. Increasing the level of structure in the prompt improves stability and peak performance.}
    \label{fig:prompt_ablation}
\end{figure}

Figure~\ref{fig:prompt_ablation} shows that the formulation of the feedback has a significant impact on training dynamics. More structured prompts, which provide richer and more constrained feedback, tend to yield slower initial performance gains but result in more stable training and higher peak performance. In particular, prompts that encourage generic, self-contained feedback (Prompts~3 and~4) appear to be especially beneficial.

\subsection{Epistemic verbalization and response length}
\label{app:epistemic_words}

\begin{figure}[h]
    \centering
    \includegraphics[width=\linewidth]{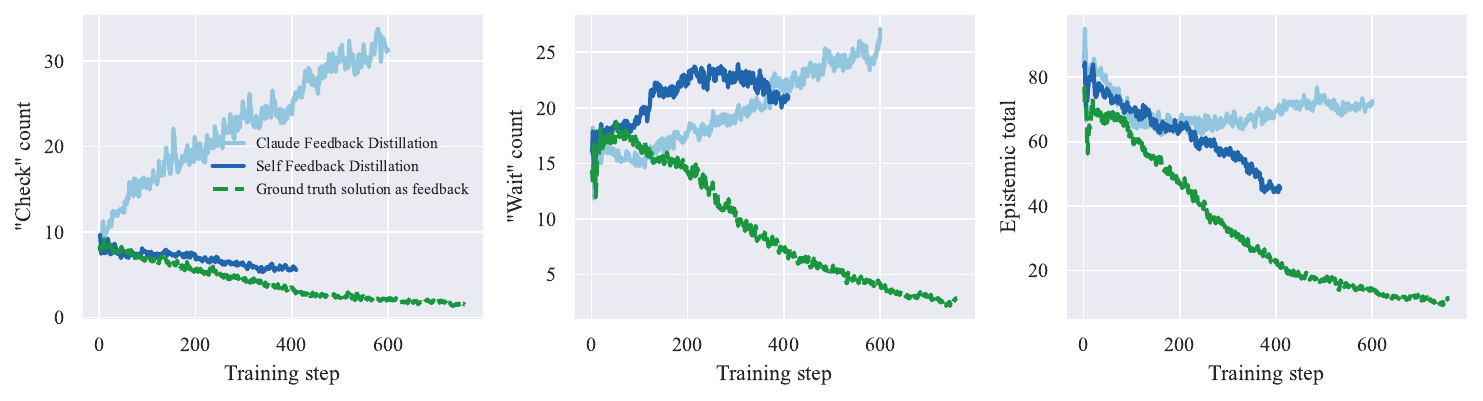}
    \caption{Occurrence counts of epistemic words (\emph{``check''}, \emph{``wait''}, and a combined epistemic total aggregating \emph{wait, hmm, perhaps, maybe, actually, alternatively, seems, might, likely, check}) over training steps for Qwen3-8B. Ground-truth feedback causes a marked decline in all three counts, whereas feedback framed as a critique by a language model — regardless of whether it comes from the student itself or a stronger model — preserves the student's epistemic verbalization.}
    \label{fig:words_count}
\end{figure}

Following \citet{kim2026understanding}, we measure epistemic verbalization by counting the average number of uncertainty words generated by the student during training.
Figure~\ref{fig:words_count} reports the per-step occurrence counts of three representative epistemic word categories (see Section~\ref{subsec:qualitative}) across training runs.
When the feedback consists of a ground-truth solution, all three counts drop sharply early in training, reflecting the suppression of epistemic verbalization.
By contrast, when feedback is framed as a critique by a language model, whether the model itself or a stronger one, the counts do not follow the same dynamic throughout training (especially when using Claude as the feedback model), indicating that it is possible to preserve the epistemic ability of the student.
We attribute this to the student mimicking the teacher's style, which is itself shaped by the feedback: a directive signal such as a worked-out solution leads the teacher to reason with greater confidence, while a critique preserves more uncertainty expression.

\begin{figure}[h]
    \centering
    \includegraphics[width=0.5\linewidth]{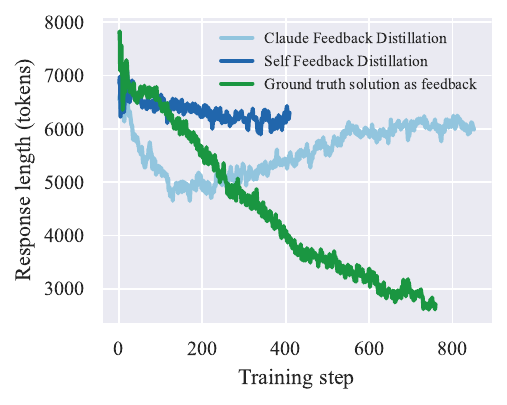}
    \caption{Qwen3-8B mean response length (in tokens) over training steps. Critique-based feedback (Claude and self-feedback) is associated with longer responses than ground-truth feedback, reflecting preserved diversity in generated trajectories.}
    \label{fig:response_length}
\end{figure}

Critique-based feedback is also associated with longer responses than ground-truth feedback (Figure~\ref{fig:response_length}), consistent with models that maintain diversity in their generated trajectories rather than converging to short, confident proofs.

\section{Feedback examples}
\label{app:feedback_examples}

The following are four representative feedback messages produced by Claude Opus 4.6 at step 100 of Qwen3-8B training with Feedback Distillation.
Each box corresponds to a different theorem-proving attempt.

\newcommand{\feedbackbox}[1]{%
  \vspace{4pt}
  \noindent\fbox{\begin{minipage}{\linewidth}
    \small #1
  \end{minipage}}
  \vspace{4pt}
}

\feedbackbox{
\begin{enumerate}
  \item When proving inequalities involving absolute values in Lean~4, use \texttt{rcases le\_or\_lt b a} and \texttt{rcases le\_or\_lt c a} to split into cases based on the sign of each expression inside the absolute values, then simplify using \texttt{abs\_of\_nonneg} and \texttt{abs\_of\_nonpos} before finishing with \texttt{linarith}.
  \item Always write and test intermediate Lean files using the \texttt{lean\_write\_file} tool rather than reasoning entirely in one's head — this ensures syntax errors and tactic failures are caught early and iteratively corrected.
  \item Structure the proof by first splitting on \texttt{le\_or\_lt (b + c) a} to handle \texttt{|a - b - c|}, then within the \texttt{a < b + c} branch, further split on the signs of \texttt{a - b} and \texttt{a - c} to reduce all absolute values to linear expressions before applying \texttt{linarith}.
\end{enumerate}
}

\feedbackbox{
\begin{enumerate}
  \item When working with natural number division in Lean~4, avoid assuming \texttt{field\_simp} will clear denominators as it does for fields; instead, use \texttt{Nat.div\_mul\_cancel} or establish divisibility (\texttt{n $\mid$ 175}) first, then rewrite to eliminate the division before algebraic manipulation.
  \item Use \texttt{omega} or \texttt{interval\_cases} after bounding $n$ (e.g., from the equation deduce $n \leq 175$) to let Lean automatically check all possible natural number values, which is often simpler than manual algebraic factoring for small search spaces.
  \item Write candidate proofs to a file using \texttt{lean\_write\_file} and check them with \texttt{lean\_check\_file} iteratively, fixing errors based on compiler output rather than reasoning abstractly about tactic behavior.
\end{enumerate}
}

\feedbackbox{
\begin{enumerate}
  \item Avoid division in Lean proofs over reals; instead, multiply through to keep all expressions polynomial, enabling \texttt{ring} and \texttt{nlinarith} to close goals without needing to prove denominators are nonzero.
  \item Derive the key intermediate fact \texttt{x * y + z * (x + y) = 0} by showing \texttt{(x - y) * (x * y + z * (x + y)) = 0} via \texttt{have : ... = x\^{}2 * (y + z) - y\^{}2 * (z + x) := by ring} combined with \texttt{linarith [h1, h2]}, then use \texttt{mul\_left\_cancel\textsubscript{0}} with \texttt{sub\_ne\_zero.mpr hxy}.
  \item After establishing \texttt{x * y + z * (x + y) = 0}, close the final goal using \texttt{nlinarith [h1, h2, sq\_nonneg x, sq\_nonneg y, sq\_nonneg z, sq\_nonneg (x - y)]} to let the nonlinear arithmetic solver combine all hypotheses without manual algebraic manipulation.
\end{enumerate}
}

\feedbackbox{
\begin{enumerate}
  \item Decompose the inequality into proving $a^2+b^2+c^2+d^2 \geq 4$ (AM-GM on squares with product $(abcd)^2=1$) and $ab+ac+ad+bc+bd+cd \geq 6$ (AM-GM on six pairwise products with product $(abcd)^3=1$), then combine with \texttt{linarith}.
  \item When using \texttt{nlinarith}, supply auxiliary lemmas as hints — e.g., \texttt{sq\_nonneg (a*b - 1)}, \texttt{sq\_nonneg (a*c - 1)}, \texttt{mul\_pos ha hb}, and the hypothesis \texttt{h} — so the solver can close the nonlinear arithmetic gap created by the constraint $abcd = 1$.
  \item Always write a complete, syntactically valid Lean file and verify it compiles using the tool before submitting; an incomplete or unreadable file scores zero regardless of mathematical correctness.
\end{enumerate}
}

\clearpage

\end{document}